%% file: main.tex
\documentclass{article}
\pdfoutput=1

\usepackage[nonatbib, final]{NeurIPS2020/neurips_2020}

\usepackage[utf8]{inputenc} 
\usepackage[T1]{fontenc}    
\usepackage{hyperref}       
\usepackage{url}            
\usepackage{booktabs}       
\usepackage{amsfonts}       
\usepackage{nicefrac}       
\usepackage{microtype}      

\usepackage{booktabs,caption}
\usepackage[flushleft]{threeparttable}

\hypersetup{
    colorlinks=true,
    linkcolor=black,
    filecolor=black,   
    citecolor=black,
    urlcolor=blue,
    pdftitle={Overleaf Example},
    pdfpagemode=FullScreen,
    }

\usepackage{graphicx} 
\usepackage{amsmath,soul}
\usepackage{amssymb}
\usepackage[capitalize]{cleveref} 

\usepackage{dsfont} 
\usepackage[font=small,skip=0pt]{subcaption} 
\usepackage{balance}
\usepackage[font=small]{caption}
\usepackage{bm} 
\usepackage{float} 
\usepackage{wrapfig} 

\title{WiSoSuper: Benchmarking Super-Resolution Methods on Wind and Solar Data}
\author{%
 Rupa Kurinchi-Vendhan \\
 California Institute of Technology \\
 \texttt{rkurinch@caltech.edu} \\
   \And
   Bj{\"o}rn L{\"u}tjens\thanks{These authors contributed equally}\\
   Massachusetts Institute of Technology\\
   \texttt{lutjens@mit.edu}
   \And
   Ritwik Gupta\footnotemark[1]\\
   University of California, Berkeley \\Defense Innovation Unit\\
   \texttt{ritwikgupta@berkeley.edu}
     \And
   Lucien Werner\footnotemark[1]\\
   California Institute of Technology \\
   \texttt{lwerner@caltech.edu}
     \And
   Dava Newman \\
   Massachusetts Institute of Technology \\
   \texttt{dnewman@media.mit.edu}
}

\begin{document}

\maketitle
\begin{abstract}
The transition to green energy grids depends on detailed wind and solar forecasts to optimize the siting and scheduling of renewable energy generation. Operational forecasts from numerical weather prediction models, however, only have a spatial resolution of $10$ to $20$-km, which leads to sub-optimal usage and development of renewable energy farms. Weather scientists have been developing super-resolution methods to increase the resolution, but often rely on simple interpolation techniques or computationally expensive differential equation-based models. Recently, machine learning-based models, specifically the physics-informed resolution-enhancing generative adversarial network (PhIREGAN), have outperformed traditional downscaling methods. We provide a thorough and extensible benchmark of leading deep learning-based super-resolution techniques, including the enhanced super-resolution generative adversarial network (ESRGAN) and an enhanced deep super-resolution (EDSR) network, on wind and solar data. We accompany the benchmark with a novel public, processed, and machine learning-ready dataset for benchmarking super-resolution methods on wind and solar data.
\end{abstract}

\input{NeurIPS2020/1_intro}

\input{NeurIPS2020/2_data}
\input{NeurIPS2020/3_results}

\input{NeurIPS2020/4_future_works}

\input{NeurIPS2020/7_acknowledgements}
\bibliographystyle{IEEEtran}
\bibliography{references}

\clearpage
\input{NeurIPS2020/6_appendix}

\end{document}

%% file: NeurIPS2020/1_intro.tex
\vspace{-3mm}
\section{Introduction}\label{sec:intro}
\vspace{-3mm}

In the United States, the national Energy Information Administration (EIA) predicts that renewable energy, predominantly wind and solar power, will contribute 42\% of the country’s electricity generation by 2050 \cite{annual_2021_2021}. To achieve this goal, operational decision-makers must integrate forecasting models into local power systems to address the spatial variability of these clean energy forms. However, current climate simulations used to obtain high-resolution data are unable to resolve the spatial characteristics necessary for accurate assessments of these energy sources in future climate scenarios, as increasing their spatial resolution is computationally expensive and provides insufficient accuracy \cite{gutowski2020ongoing}. Numerical weather predictions (NWPs) provide short-term climatological forecasting data at a horizontal resolution of 10 to 20-km \cite{sweeney2011adaptive, clifton2018wind, diagne2013review}, while energy planning requires wind and solar data at a smaller, more local scale, on the order of 2-km \cite{osti_1427970}.


In the field of computer vision, researchers enhance the resolution of a data field through single-image super-resolution \cite{yang2014single}. Although the problem of super-resolution is inherently ill-posed---coarsened low-resolution (LR) input data can map to infinitely many high-resolution (HR) outputs---machine learning-based approaches offer a cost-effective and accurate method of generating the high-resolution data needed to predict the effect climate has on power generation \cite{bano2020configuration, Watson_2020, Leinonen_2021}.

Super-resolution techniques have increasingly been applied to climate data in recent years \cite{cheng2020reslap, Watson_2020, vandal2017deepsd, kaltenboeck2012image}. As these methods have the potential to vastly improve spatial SR for wind and solar predictions, it is important to rigorously validate them to ensure that their SR outputs are accurate and realistic. 

In this work we contribute:
\vspace{-1pt}
\begin{itemize}
    \item an extensible benchmark for determining accurate and physically consistent super-resolution models for wind and solar data; 
    \item a novel application of state-of-the-art convolutional neural network (CNN)- and generative adversarial network (GAN)-based super-resolution techniques to a task from the physical sciences;
    \item and a novel publicly available machine learning-ready dataset for the super-resolution of wind speeds and solar irradiance fields.
\end{itemize}

%% file: NeurIPS2020/2_data.tex

\section{Data}
\vspace{-3mm}

The training data was obtained from the National Renewable Energy Laboratory's (NREL's) Wind Integration National Database (WIND) Toolkit and the National Solar Radiation Database (NSRDB), with a focus on the continental United States. Wind velocity data is comprised of westward (ua) and southward (va) wind components, calculated from wind speeds and directions 100-km from Earth's surface. The WIND Toolkit has a spatial resolution of $2\text{km}\times1\text{hr}$ spatiotemporal resolution \cite{WIND, draxl_clifton_hodge_mccaa_2015, WIND2, WIND3}. Our wind dataset contains data sampled at a 4-hourly temporal resolution for the years 2007 to 2013. Wind test data are sampled at a 4-hourly temporal resolution for the year 2014. 

Additionally, we consider solar irradiance data from the NSRDB in terms of direct normal irradiance (DNI) and diffused horizontal irradiance (DHI) at an approximately $4\text{km}\times1/2\text{hr}$ spatiotemporal resolution \cite{sengupta2018national}. The solar dataset produced for this work samples data at an hourly temporal resolution from  6 am to 6 pm for the years 2007 to 2013. Solar test data are sampled at an hourly temporal resolution from 6 am to 6 pm for the years 2014 to 2018. More information about the wind and solar datasets is available in the Appendix.

\section{Models}
\vspace{-3mm}
We examined five super-resolution techniques---PhIREGAN \cite{Stengel16805}, ESRGAN  \cite{wang2018esrgan}, EDSR  \cite{lim2017enhanced}, SR CNN, and bicubic interpolation---on their ability to perform 5$\times$ super-resolution, e.g., 10-km to 2-km spatial resolution, on wind and solar data fields. The PhIREGAN is structured in a two-phase process: an LR to medium-resolution (MR) step and an MR to HR step. In this work, we examine the MR → HR step of this model for both wind and solar data, which corresponds to a deep fully convolutional neural network based on the super-resolution generative adversarial network (SRGAN) \cite{8099502}. Like the PhIREGAN, ESRGAN improves upon SRGAN's network. The ESRGAN has been shown to reproduce the simulated power spectral density of near-surface winds \cite{singhnumerical, manepalli}, which encourages its application to climate data. The EDSR model has an architecture similar to that of the SRResNet \cite{8099502}, with simplifications to preserve high-frequency features that would have otherwise been blurred. We modify the upsampling block of the ESRGAN and the EDSR to be compatible with 5$\times$ super-resolution. We included the pre-trained network of the PhIREGAN---the SR CNN (distinct from the SRCNN \cite{dong2015image}), which minimizes content loss---and bicubic interpolation as baselines for this benchmark.

%% file: NeurIPS2020/3_results.tex
\section{Benchmarking Results}\label{sec:results}
\vspace{-3mm}

\begin{figure}[h]
    \centering
    \includegraphics[width = \textwidth]{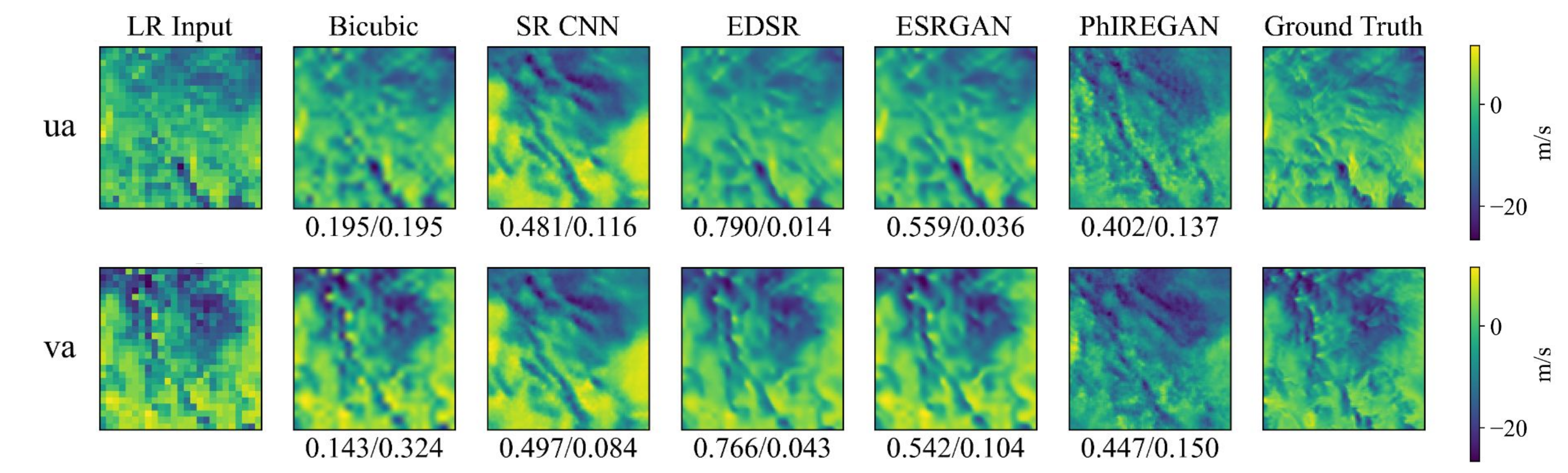}
    \caption{A comparison of wind outputs from each model, with reported SSIM/MSE values. While the outputs of current models are pixelated (bicubic), blurry (SR CNN), or speckled (PhIREGAN), our contributed implementations (EDSR, ESRGAN) are the most accurate.}
    \label{fig:wind_sample}
\end{figure}

\begin{figure}[h]
    \centering
    \includegraphics[width = \textwidth]{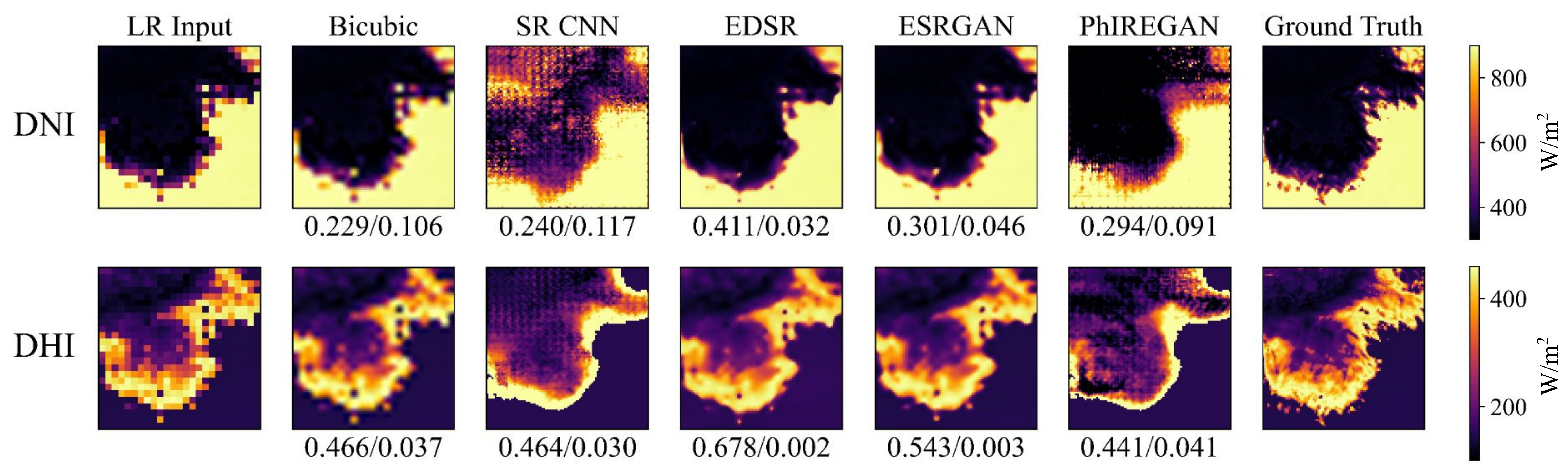}
    \caption{A comparison of solar outputs from each model, with reported SSIM/MSE values. EDSR and the ESRGAN outperform other models in terms of accuracy. The PhIREGAN and SR CNN have gridded artifacts, but sharper structures.}
    \label{fig:solar_sample}
\end{figure}

Figures \ref{fig:wind_sample} and \ref{fig:solar_sample} compare sample outputs from each model. Qualitatively, the results in Figures \ref{fig:wind_sample} and \ref{fig:solar_sample} show that deep learning models---most noticeably the PhIREGAN---produce outputs with sharper structures and small-scale details. This is a result of the fact that models such as EDSR that use an L1 loss function do not account for perceptual or adversarial loss like the ESRGAN and PhIREGAN do. Enlarged versions of the images in Figures \ref{fig:wind_sample} and \ref{fig:solar_sample} are provided in Figures \ref{fig:ua_wind_sample}-\ref{fig:dhi_solar_sample} in the Appendix.

To assess the ability of each model to accurately recreate the ground truth data, we applied image quality metrics such as the peak signal-to-noise ratio (PSNR), structural similarity index (SSIM), relative mean-squared error (MSE), and mean absolute error (MAE). Table \ref{table:average_metric_values} shows the average values of each metric across all test data, for both wind and solar data fields. All deep learning methods assessed outperform bicubic interpolation. Additionally, models which are trained to minimize content loss, such as the PhIREGAN's pre-trained CNN and EDSR have stronger metric values than the PhIREGAN, which is trained to minimize adversarial loss. EDSR outperforms all other models on image quality metrics for both wind and solar data, followed by ESRGAN.

\begin{table}
    \centering
    \renewcommand{\arraystretch}{1.15}
    \begin{tabular}{ |l|c|c|c|c|c|c|c|c|  }
         \cline{2-9}    
         \multicolumn{1}{c|}{} & \multicolumn{4}{|c|}{Wind} & \multicolumn{4}{|c|}{Solar}\\
         \hline
         Model & PSNR$\uparrow$ & SSIM$\uparrow$ & MSE$\downarrow$ & MAE$\downarrow$ & PSNR$\uparrow$ & SSIM$\uparrow$ & MSE$\downarrow$ & MAE$\downarrow$\\
         \hline
         PhIREGAN & {29.11} & {0.46} & {0.12} & {0.27} & {28.85} & {0.44} & {0.34} & {0.56}\\
         EDSR & \textbf{32.25} & \textbf{0.83} & \textbf{0.02} & \textbf{0.09} & \textbf{32.88} & \textbf{0.63} & \textbf{0.29} & \textbf{0.19}\\
         ESRGAN & \textit{30.96} & \textit{0.69} & \textit{0.03} & \textit{0.12} & \textit{31.82} & \textit{0.56} & \textit{0.31} & \textit{0.25}\\
         SR CNN & 28.94 & 0.52 & 0.15 & 0.31 & 29.05 & 0.52 & 0.33 & 0.42\\
         Bicubic & 28.83 & 0.36 & 0.21 & 0.36 & 28.78 & 0.40 & 0.46 & 0.61\\
         \hline
    \end{tabular}
    \vspace{0.3cm}
    \caption{Summary of Average Metric Values. The models we introduce to this field (EDSR \& ESRGAN) are most accurate across all metrics.}
    \label{table:average_metric_values}
    \vspace{-0.8cm}
\end{table}

Image quality metrics are limited in scope and do not offer a comprehensive method of evaluating wind and solar data fields as they cannot capture the ability to replicate high-frequency features. We validated the super-resolved wind speed outputs by generating kinetic energy spectra for each model which measure the distribution of energy across the various wavenumbers, $k$ \cite{kolmogorov1991local, kolmogorov1941dissipation}.

\begin{figure}[h]
    \vspace{-0.3cm}
    \centering
    \includegraphics[width=0.6\textwidth]{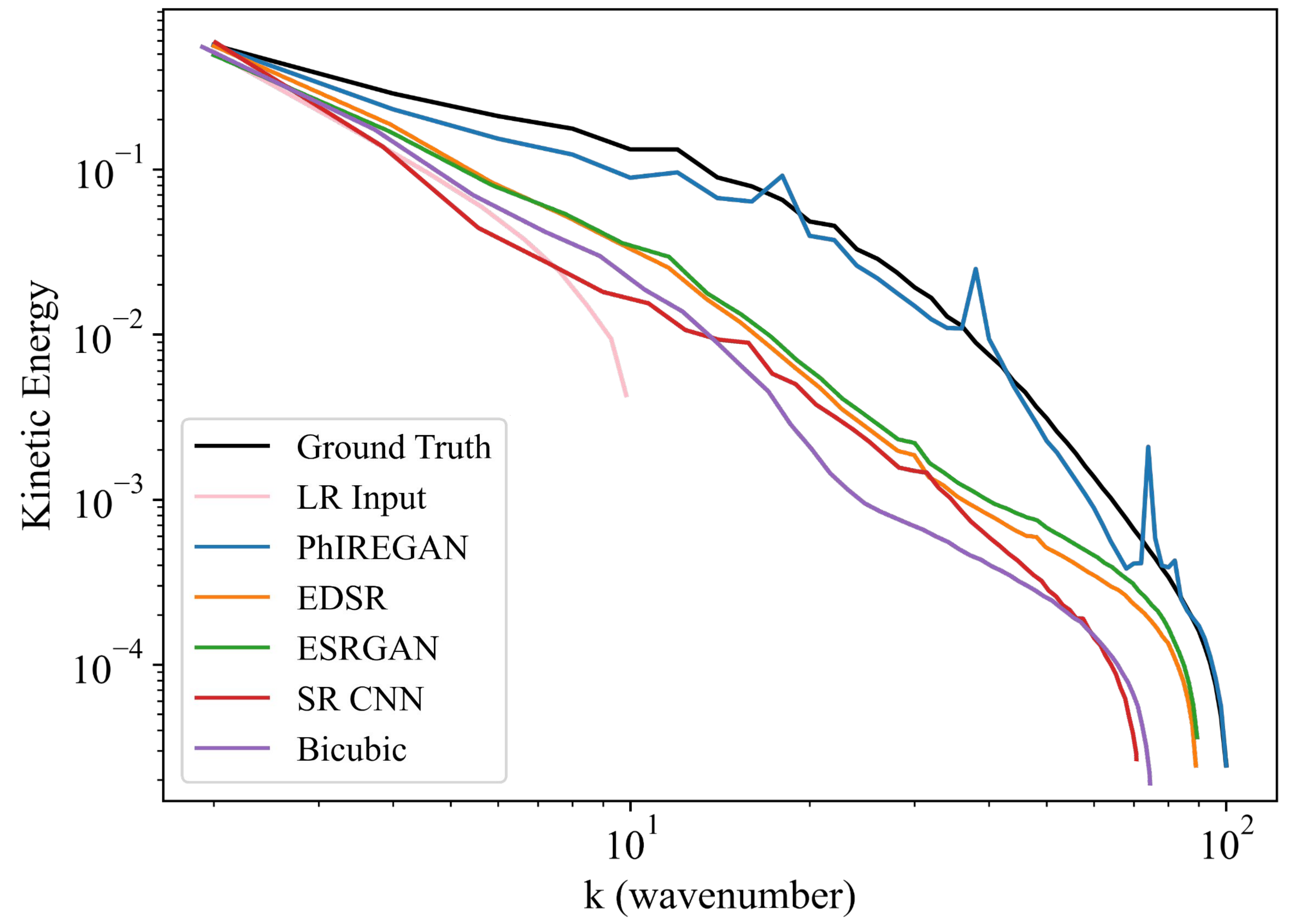}
    \caption{Kinetic energy spectra for the data fields corresponding to each HR output, averaged over all wind test data. The PhIREGAN most closely matches the turbulent physics of the ground truth data.}
    \label{fig:energy}
    \vspace{-0.3cm}
\end{figure}

Consistent with the perception-distortion trade-off \cite{blau2018perception}, although the downscaled wind maps generated by the PhIREGAN performed slightly worse on accuracy metrics, they outperformed other HR outputs in terms of visual and physical fidelity. In Figure \ref{fig:energy}, energy is conserved in the inertial range of each energy spectrum and cascades at higher wavenumbers. Bicubic interpolation and the SR CNN, which have stronger metric values, perform visibly worse than the PhIREGAN in capturing high-frequency data consistent with turbulence theory. The EDSR and ESRGAN outputs remain fairly consistent with the ground truth data especially at larger wavenumbers, despite performing optimally on the image quality metrics. As seen in Figure \ref{fig:energy}, both PhIREGAN and ESRGAN most closely match the energy spectrum of the ground truth data, which suggests that GAN-based wind downscaling approaches most successfully learn physical relationships across various frequencies that classical techniques are less capable of recreating.

For solar data, we compare super-resolved outputs by generating normalized semivariograms that show the directionally-averaged spatial autocorrelation of gradients in an image as a function of a radius $r$ \cite{matheron1963principles}. 

\begin{figure}[h]
    \vspace{-0.3cm}
    \centering
    \includegraphics[width = \textwidth]{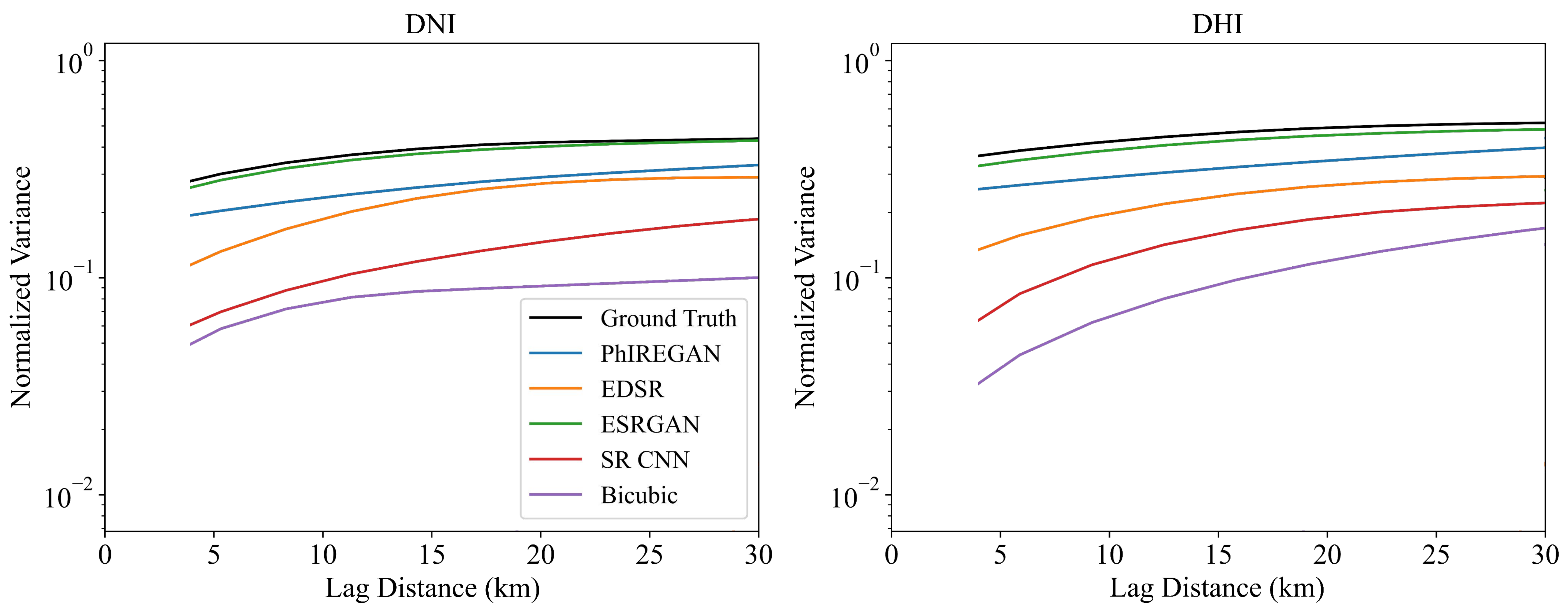}
    \caption{Normalized semivariograms for each solar output. The ESRGAN most closely matches the semivariance of the ground truth data for both DNI and DHI. This means that ESRGAN best captures the spatial correlations between pixels.}
    \label{fig:semivariograms}
    \vspace*{-0.3cm}
\end{figure}

Figure \ref{fig:semivariograms} focus on the $4$-km $< r < 20$-km regime where each model generates high-frequency data. GAN-based models such as the PhIREGAN and ESRGAN outperform other methods on generating spectrally-consistent solar data. In both the DNI and DHI semivariograms, the ESRGAN has the lowest deviation from the ground truth data, visibly outperforming other models.

%% file: NeurIPS2020/4_future_works.tex
\section{Discussion and Future Works}
\vspace{-2.5mm}

These results indicate that the perception-distortion trade-off holds for the super-resolution of wind and solar data. Our benchmark shows that CNN-based image-processing techniques (e.g., EDSR) are likely to achieve higher image similarity (in MAE, RMSE, PSNR, SSIM) while GAN-based methods achieve higher spectral similarity (see \Cref{fig:energy,fig:semivariograms}). We show that GAN-based models have significant applications in climate scenarios, as they most reliably generate results that match the spectral dynamics of the ground truth. Of the two GAN-based models benchmarked, ESRGAN performs best in super-resolving solar data whereas PhIREGAN performs best for wind data.

Super-resolution is an ill-posed problem for which a single low-resolution climate scenario can correspond to multiple high-resolution scenarios. In future works, we will extend the benchmark to stochastic super-resolution techniques such as variational auto-encoders (VAEs) \cite{Kingma_2014}, normalizing flows \cite{Lugmayr_2020}, and diffusion-based \cite{Li_2021a} models and metrics such as CRPS \cite{Leinonen_2021}. Validating these models on data outside of NREL's WIND Toolkit and NSRDB will also widen the scope of this study and enable in-depth model comparisons.

\section{Conclusion}
\vspace{-2.5mm}

In this paper, several state-of-the-art super-resolution methods are thoroughly evaluated on national wind and solar data. Comprehensive experimental results demonstrate how these methods perform with respect to accuracy and plausibility. The benchmarking assessments show the qualitative and quantitative performance and limitations of each model. Our GitHub repository\footnote{\href{https://github.com/RupaKurinchiVendhan/WiSoSuper}{https://github.com/RupaKurinchiVendhan/WiSoSuper}} provides a machine learning-ready dataset from NREL's WIND Toolkit and NSRDB and detailed instructions for generating a new one, compiled implementations of each model examined in this work, and an accessible platform for assessing the turbulent flow and spatial autocorrelation of any super-resolved wind and solar output.

%% file: NeurIPS2020/7_acknowledgements.tex

\subsection*{Acknowledgements}
We would like to thank the Caltech Student-Faculty Programs office and Dr. Steven Low's Netlab for funding this work. We gratefully acknowledge the computational support from Microsoft AI for Earth. We would also like to thank Karen Stengel and Michael Rossol for their assistance.

Data is obtained from the U.S. Department of Energy (DOE)/NREL/ALLIANCE.

%% file: NeurIPS2020/6_appendix.tex
\section{Appendix}
\textbf{Ethical Considerations}\label{sec:appendix_ethics}

This work aims to increase the efficiency of integrating renewable energy forms in US power systems. Since the focus of this study is on the continental US, our results may not be generalizable to other nations and geographic regions. However, we provide a replicable codebase and dataset that will enable transition of our results to other national wind and solar datasets.

\textbf{Dataset}\label{sec:appendix_dataset}

The data that will be available for download upon publication consists of separate datasets for wind and solar. We transform 2D data arrays of wind speed and direction into  corresponding ua and va wind speed components. These are chipped into 100$\times$100 patches. Low resolution imagery is obtained by sampling high resolution data at every fifth data point as instructed by NREL's guidelines.

The NSRDB database is formatted differently from the NREL database. A 1D array of data points is provided along with latitude and longitude metadata for each point. We re-arrange this 1D array into a 2D image based on the lat/long metadata.

All data files are made available as PNGs in their respective LR or HR resolutions. The code for for retrieving and processing WIND Toolkit and NSRDB data are accessible in the code library. Additional specifications of the wind and solar data are summarized in Table \ref{table:datasets}.

\begin{table}[h]
    \renewcommand{\arraystretch}{1.15}
    \centering
    \begin{tabular}{ |l||c|c|  }
     \hline
     Data & Wind & Solar\\
     \hline
     Institute & NREL & NREL\\
     Model & WIND Toolkit & NSRDB\\
     Spatial Resolution & 2 km &  4 km\\
     Temporal Resolution & 4-hr & 1-hr\\
     Years & 2007-2013 &  2007-2013\\
     Number of Files & 153,600 & 153,600\\
     HR Dimensions & 100$\times$100 & 100$\times$100\\
     LR Dimensions & 20$\times$20 & 20$\times$20\\
     Colormap & Viridis & Inferno\\
     \hline
    \end{tabular}
    \vspace{0.3cm}
    \captionof{table}{Wind and solar data specifications.}
    \label{table:datasets}
\end{table}

\begin{figure}[h]
    \centering
    \includegraphics[width = \textwidth]{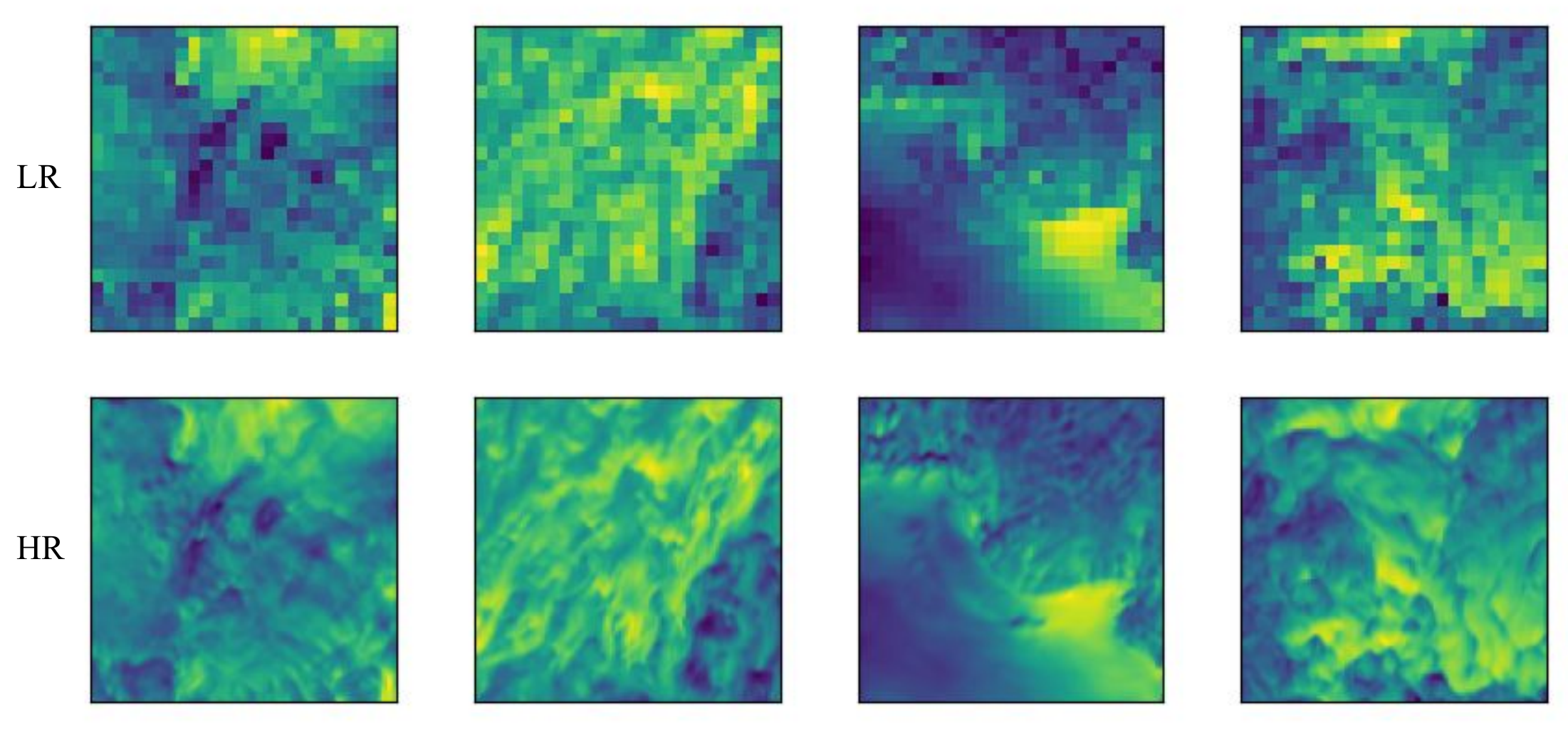}
    \caption{The LR and HR pairs from the generated wind dataset, with the coarsened data being 5$\times$ downsampling.}
    \label{fig:wind_data}
\end{figure}

\begin{figure}[h]
    \centering
    \includegraphics[width =\textwidth]{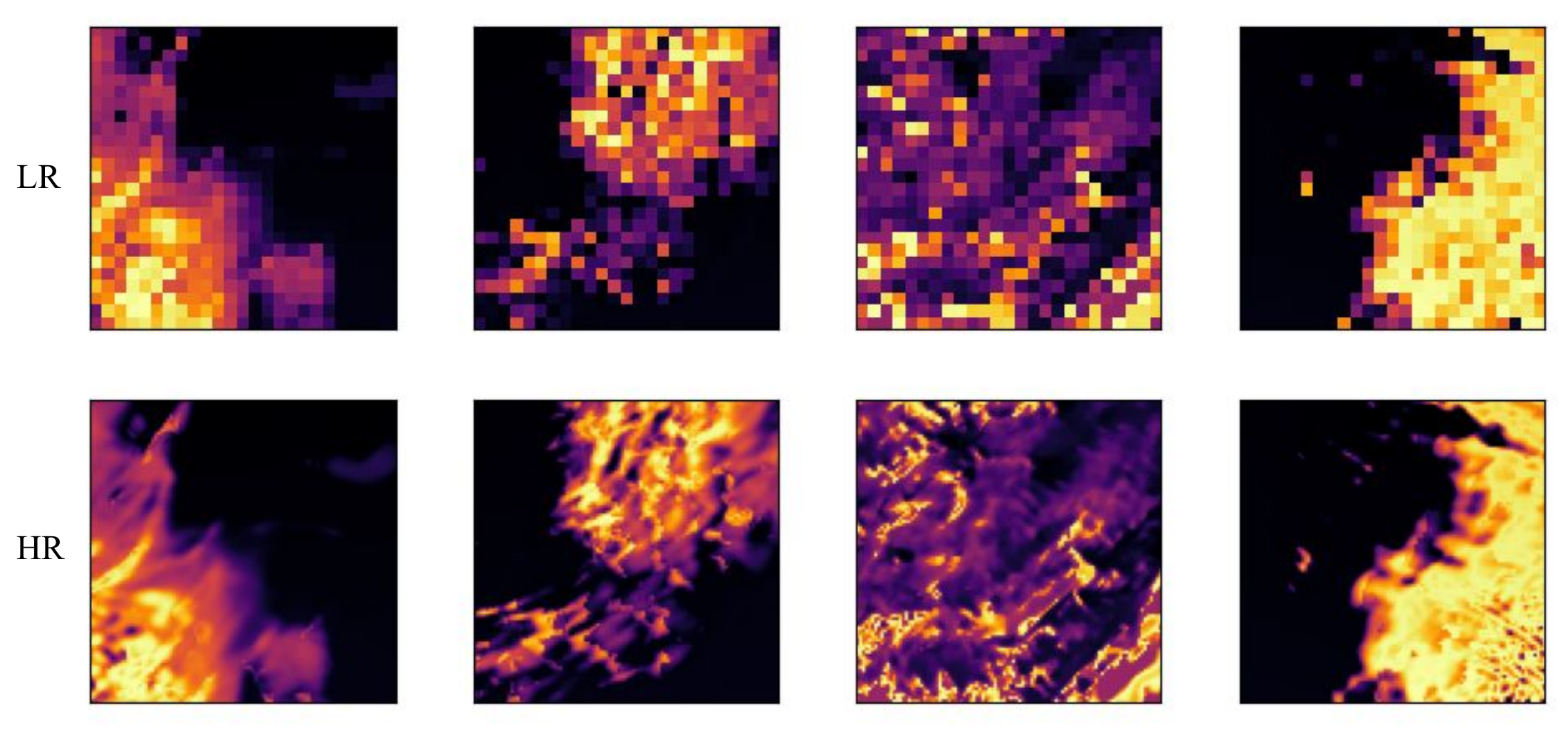}
    \caption{The LR and HR pairs from the generated solar dataset, with the coarsened data being 5$\times$ downsampling.}
    \label{fig:solar_data}
\end{figure}

Examples of LR and HR imagery for both wind and solar datasets are visualized in Figures \ref{fig:wind_data} and \ref{fig:solar_data}.

\textbf{Training}\label{sec:appendix_models}

The following training hyperparameters were used:
\begin{itemize}
    \item The PhIREGAN model was trained on NREL's Eagle computing system (the pre-trained model weights were used to run inference for this project ESRGAN was pre-trained for 20 epochs and GAN-trained for an additional 20 epochs.
    \item The EDSR network was trained for 20 epochs.
    \item The ESRGAN was pre-trained for 20 epochs and GAN-trained for an additional 20 epochs.
The implementations offered in the GitHub repository for this project default to these training settings.
\end{itemize}

\textbf{Supplementary Figures}

\begin{figure}[h]
    \centering
    \includegraphics[width = 0.9\textwidth]{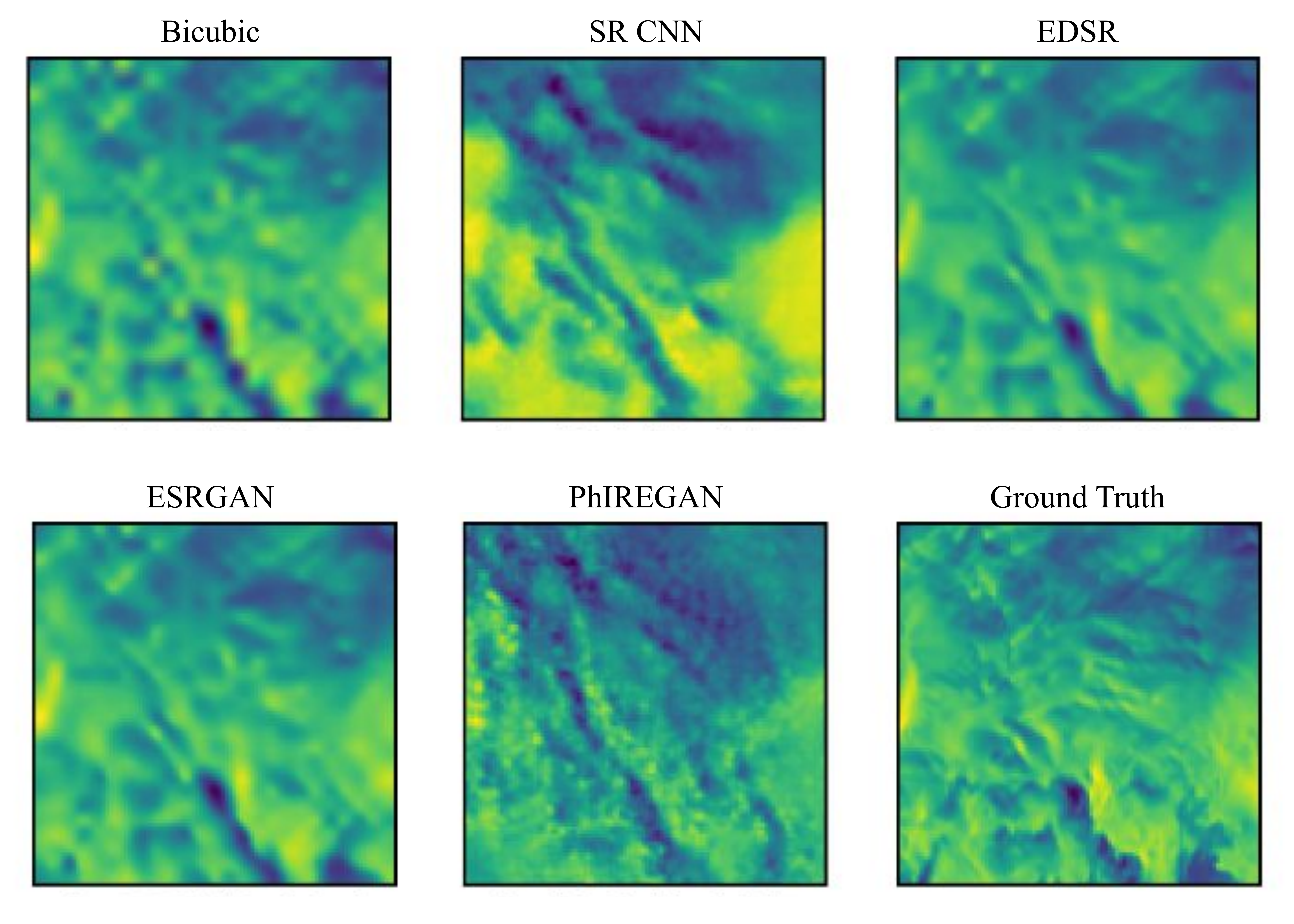}
    \caption{The ua SR outputs from each of the five models examined in this work, on a larger scale for ease of comparison.}
    \label{fig:ua_wind_sample}
\end{figure}

\begin{figure}[h]
    \centering
    \includegraphics[width = 0.9\textwidth]{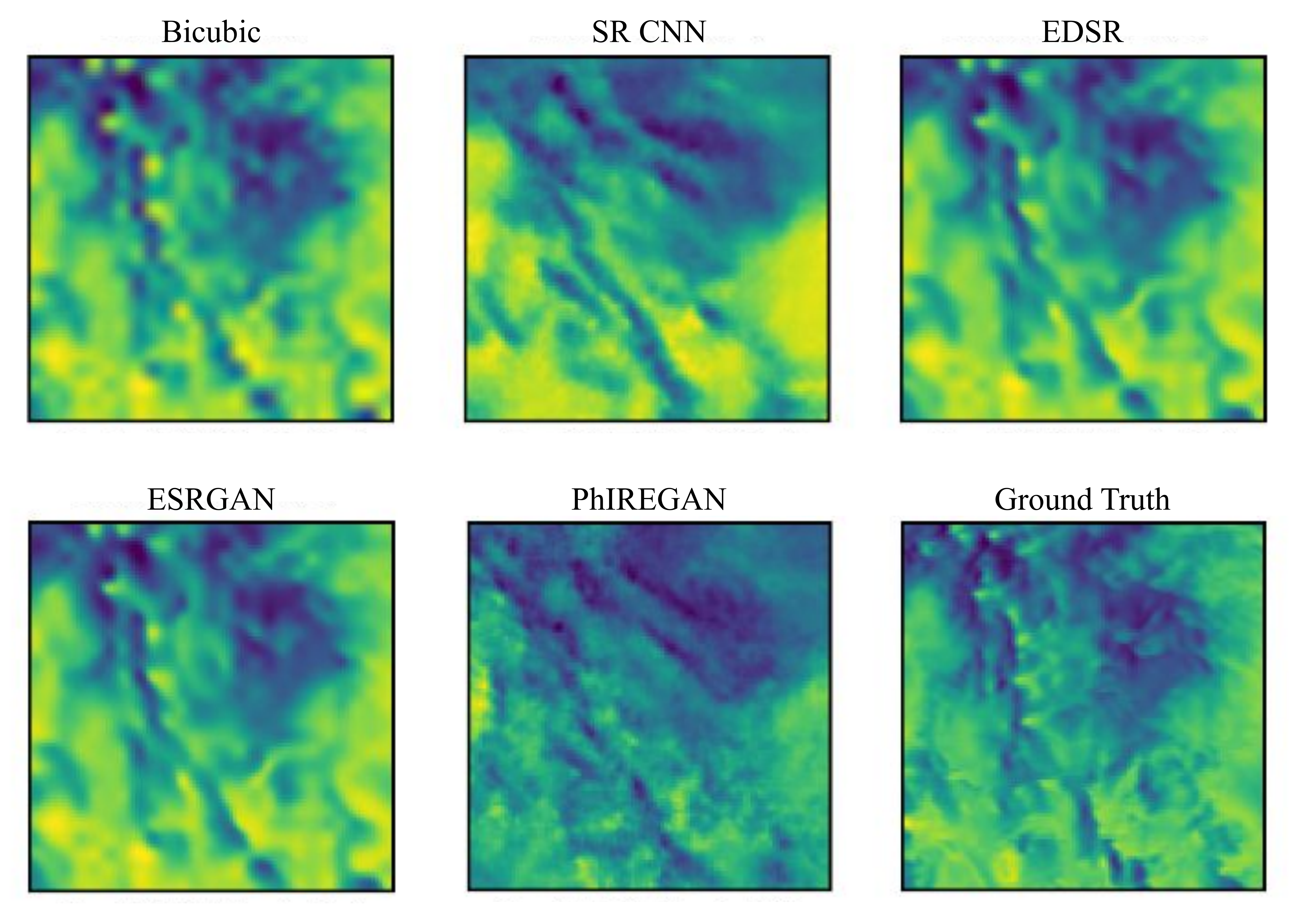}
    \caption{The va SR outputs from each of the five models examined in this work, on a larger scale for ease of comparison.}
    \label{fig:va_wind_sample}
\end{figure}

\begin{figure}[]
    \centering
    \includegraphics[width = 0.9\textwidth]{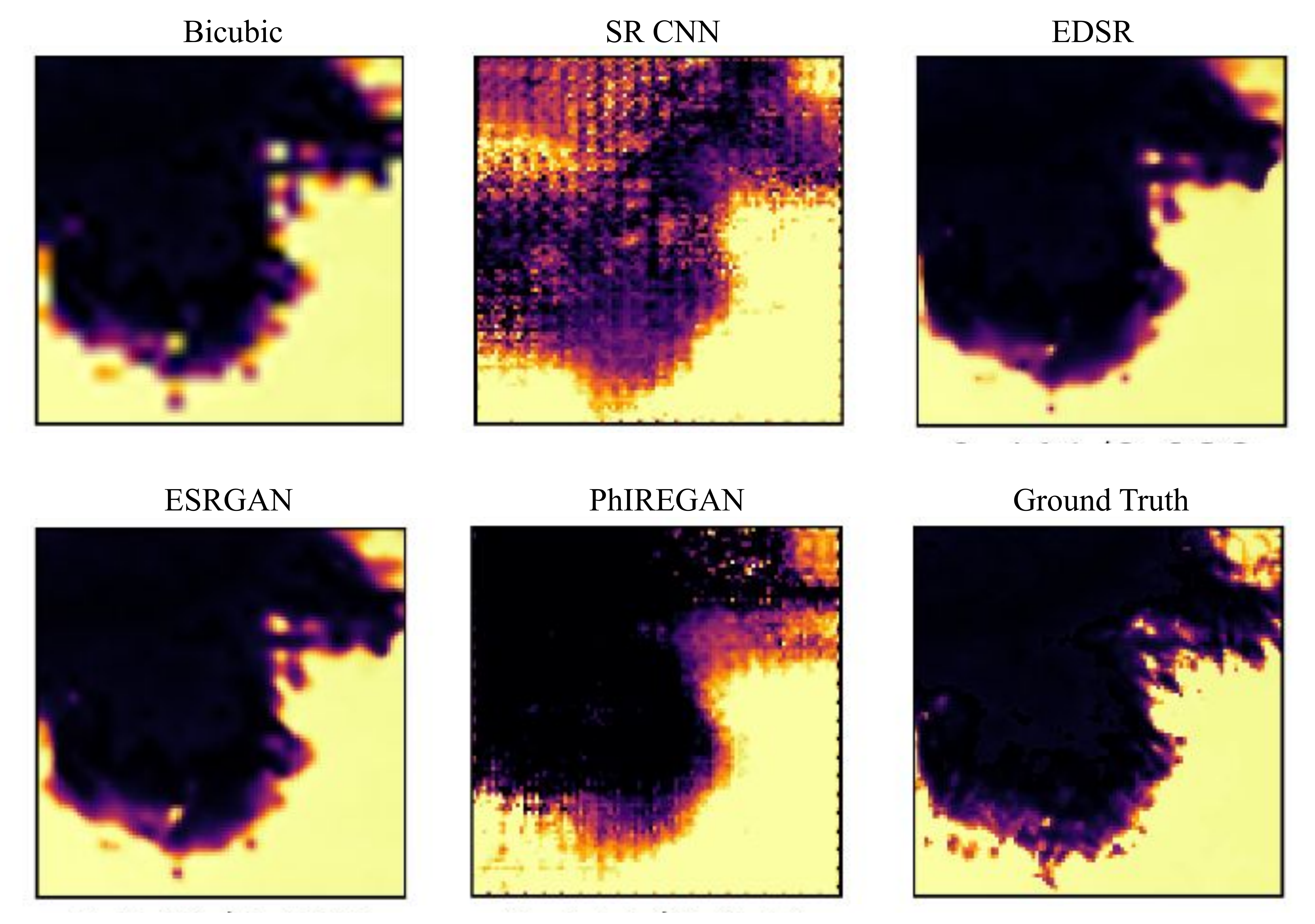}
    \caption{The DNI SR outputs from each of the five models examined in this work, on a larger scale for ease of comparison.}
    \label{fig:dni_solar_sample}
\end{figure}

\begin{figure}[h]
    \centering
    \includegraphics[width = 0.9\textwidth]{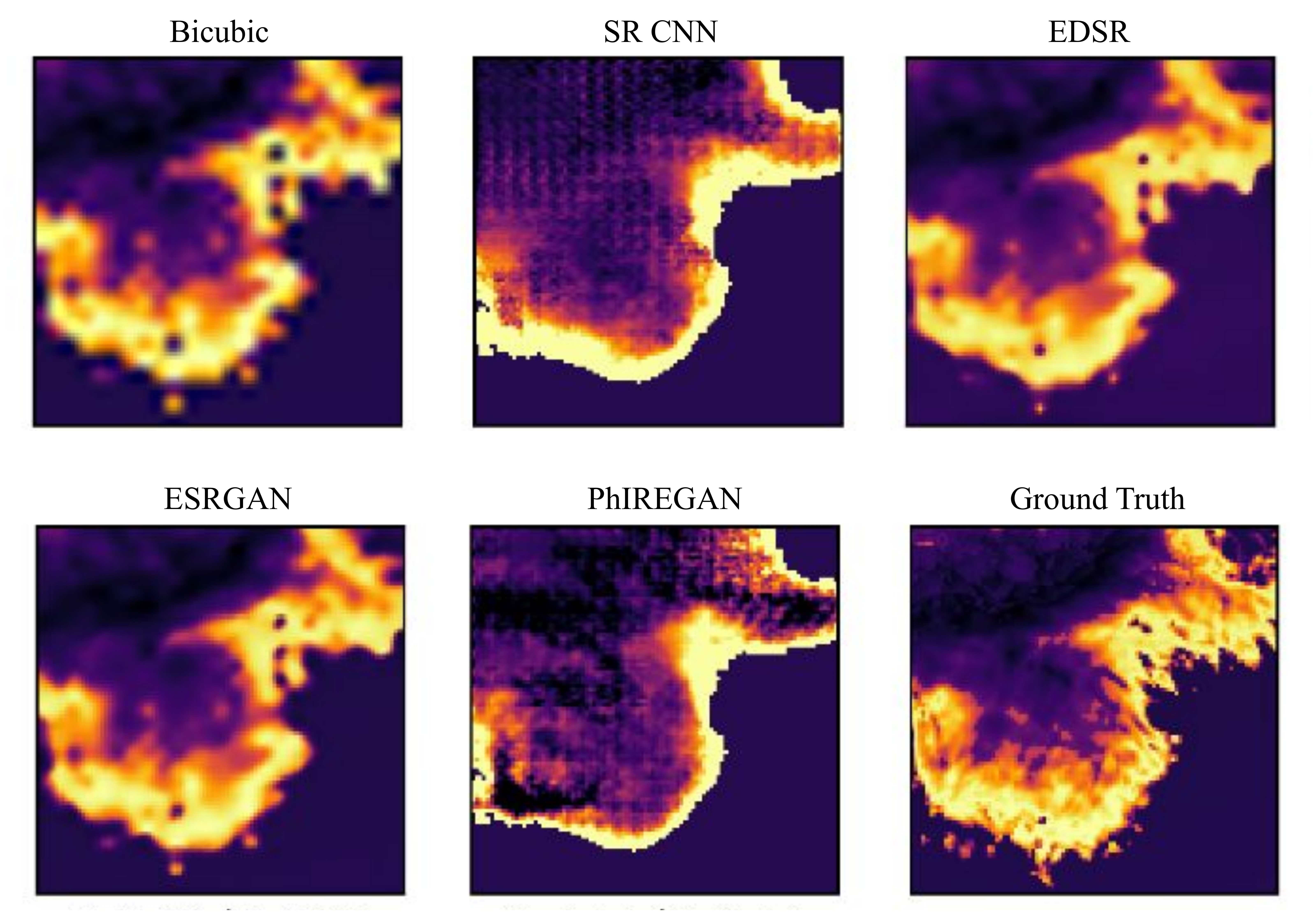}
    \caption{The DHI SR outputs from each of the five models examined in this work, on a larger scale for ease of comparison.}
    \label{fig:dhi_solar_sample}
\end{figure}

In support of the qualitative discussion of sample SR outputs in Section \ref{sec:results}, Figures \ref{fig:ua_wind_sample}-\ref{fig:dhi_solar_sample} provide the same images at a larger scale for better viewing and comparison.

\begin{figure}[h]
    \centering
    \includegraphics[width = 0.9\textwidth]{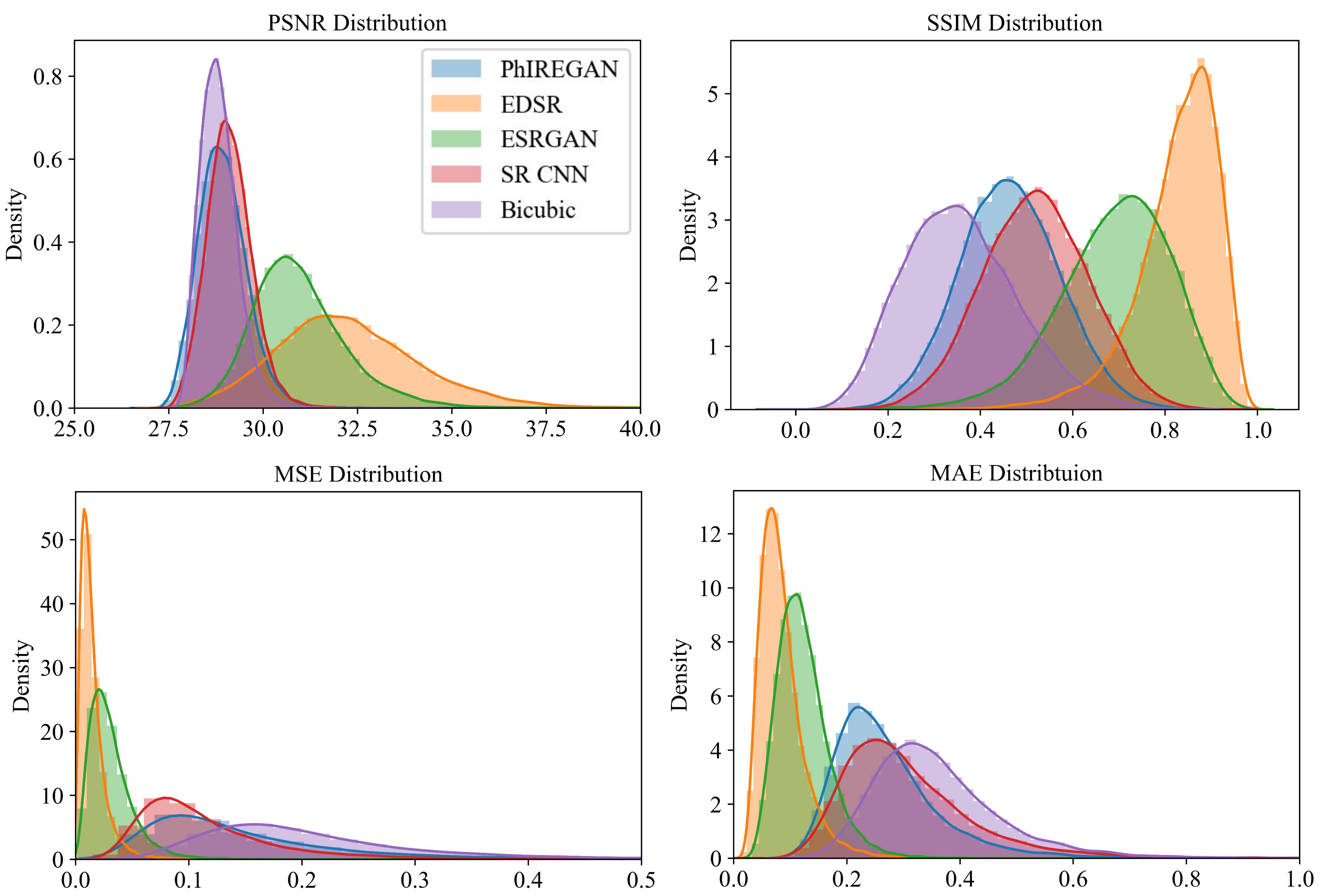}
    \caption{The distribution of each distortion metric across all wind test data, averaged over both ua and va.}
    \label{fig:wind_distribution}
\end{figure}

\begin{figure}[h]
    \centering
    \includegraphics[width = 0.9\textwidth]{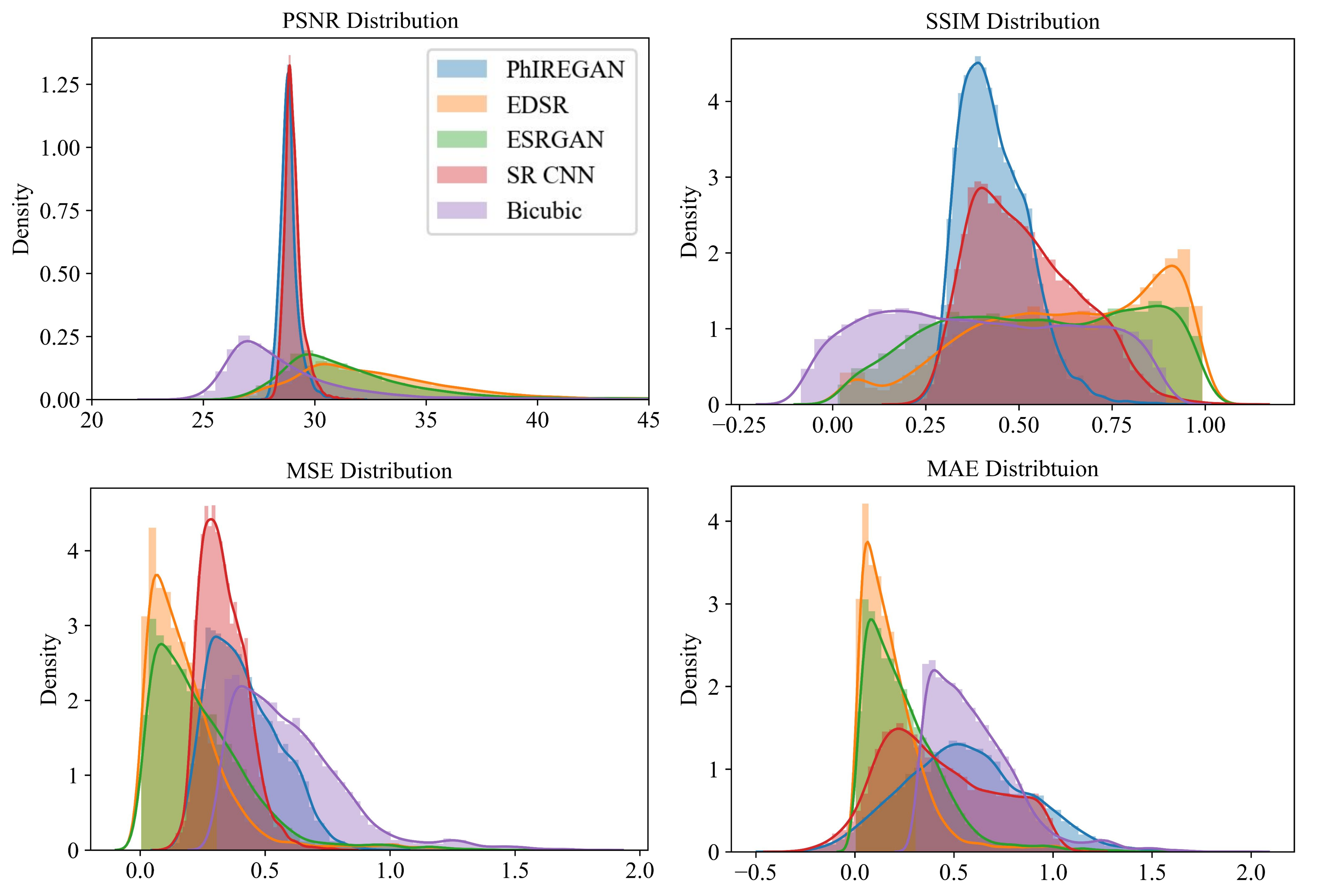}
    \caption{The distribution of each distortion metric across all solar test data, averaged over both DNI and DHI.}
    \label{fig:solar_distribution}
\end{figure}

Figures \ref{fig:wind_distribution} and \ref{fig:solar_distribution} show the distributions of each image quality metric examined in this work across test wind and solar data, respectively.

